\documentclass[conference]{IEEEtran}
\IEEEoverridecommandlockouts
\usepackage{cite}
\usepackage{amsmath,amssymb,amsfonts}
\usepackage{algorithmic}
\usepackage{algorithm}
\usepackage{graphicx}
\usepackage{textcomp}
\usepackage{xcolor}
\def\BibTeX{{\rm B\kern-.05em{\sc i\kern-.025em b}\kern-.08em
    T\kern-.1667em\lower.7ex\hbox{E}\kern-.125emX}}

\begin{document}

\title{Adaptive Structural Hyper-Parameter Configuration by Q-Learning}

\author{\IEEEauthorblockN{Haotian Zhang, Jianyong Sun and Zongben Xu}
\IEEEauthorblockA{\textit{ School of Mathematics and Statistics,} \\
\textit{National Engineering Laboratory for Big Data Analytics,}\\
\textit{Xi'an Jiaotong University, Xi'an, China}\\
zht570795275@stu.xjtu.edu.cn, jy.sun@xjtu.edu.cn, zb.xu@xjtu.edu.cn}}

\maketitle

\begin{abstract}
Tuning hyper-parameters for evolutionary algorithms is an important issue in computational intelligence. Performance of an evolutionary algorithm depends not only on its operation strategy design, but also on its hyper-parameters. Hyper-parameters can be categorized in two dimensions as structural/numerical and time-invariant/time-variant. Particularly, structural hyper-parameters in existing studies are usually tuned in advance for time-invariant parameters, or with hand-crafted scheduling for time-invariant parameters. In this paper, we make the first attempt to model the tuning of structural hyper-parameters as a reinforcement learning problem, and present to tune the structural hyper-parameter which controls computational resource allocation in the CEC 2018 winner algorithm by Q-learning. Experimental results show favorably against the winner algorithm on the CEC 2018 test functions.
\end{abstract}



\section{Introduction}

Evolutionary algorithm (EA) is an important research area in computation intelligence. Over several decades, fruitful research studies have been conducted. Example EAs, such as differential evolution (DE)~\cite{Price1999Introduction,Das2011Differential}, particle swarm optimization (PSO)~\cite{Russell1995New}, CMA-ES~\cite{Hansen2001Completely} and many others, have attracted a great amount of attentions.

The hybridization of EAs has also achieved great success, such as DE/EDA~\cite{sun2005de,Sun2011TwostageEA},  SaDE~\cite{Qin2009Differential}, jSO~\cite{Brest2017Single} and others. The aim of hybrid EAs is to take advantages of the pros of different EAs and to compensate the cons of these EAs for favorable algorithmic performance. Very recently, the combination of univariate sampling and CMA-ES~\cite{Hansen2001Completely}, \cite{Hansen2003Reducing}, called  HSES~\cite{Geng2018hygrid}, has achieved the best performance for the CEC 2018 test functions.

\begin{figure}
\centerline{\includegraphics[width=1\columnwidth]{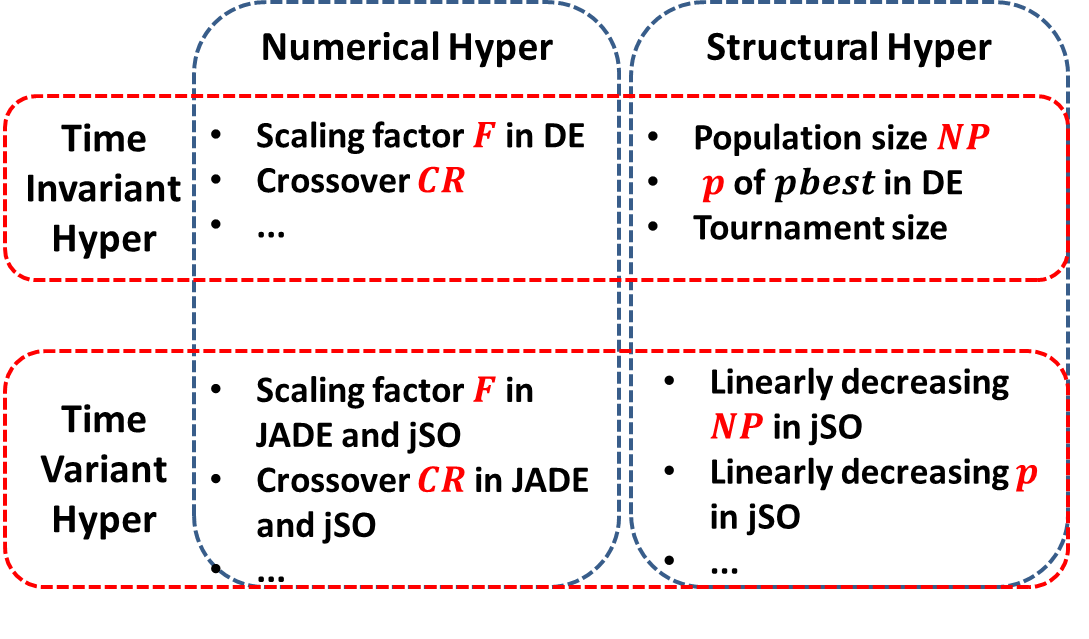}}
\caption{Categories of hyper-parameters in EAs.}\label{EAhyper}
\end{figure}

In all the developed EAs, these always exist more or less hyper-parameters. Those hyper-parameters can be categorized in two dimensions as shown in Fig.~\ref{EAhyper}. In one hand, the hyper-parameters can be time variant or invariant. For instances, the scaling factor $F$ and crossover rate $CR$ can be either fixed like in traditional DE, or adaptively changed such as in JADE~\cite{Zhang2009JADE}. Since the scaling factor and crossover rate are directly responsible for creating new solutions through arithmetic and/or logic operations, they are also categorized as ``numerical hyper-parameters". On the other hand, for hyper-parameters such as the population size, the integer $p$ in the current-to-best DE operator~\cite{Brest2017Single}, the tournament size and others, are categorized as ``structural hyper-parameters" since they do not directly involve in the solution creation procedure.

The performance of an EA depends not only on its core components including recombination and selection operations, but also on its hyper-parameters. Tuning hyper-parameter for optimal algorithmic performance can be cumbersome, time-consuming and tedious. What's worse, hybrid algorithms can bring extra structural hyper-parameters. For example, they will require not only the tuning of each composing algorithm's parameters, but also need to control the resource allocation for each of them.

Taking the winner of CEC 2018, HSES~\cite{Geng2018hygrid}, as an example, two heuristics including a univariate sampling algorithm and CMA-ES are carried out sequentially. In HSES, except the numerical hyper-parameters used in univariate sampling and CMA-ES, the allocation of resources, or precisely the number of iterations $K$ used by the first univariate sampling, is an important structural hyper-parameter. It could significantly influence the performance of HSES.

Algorithms such as Bayesian optimization (BOA)~\cite{Frazier2018A}, sequential model-based optimization for algorithm configuration (SMAC)~\cite{hutter2011sequential} and others, can be applied on tuning time-invariant numerical hyper-parameter. For examples, BOA has been successfully applied to tune numerical hyper-parameters in~\cite{roman2016bayesian}\cite{huang2019automatic}. Interested readers please see~\cite{huang2019a} for a review of the tuning techniques.

In these methods, tuning time-invariant numerical hyper-parameters is modeled as optimizing black-box optimization problem. Once optimized, the resultant hyper-parameters are fixed during the optimization procedure. However, for time-variant hyper-parameters, we cannot use these methods simply because of the time-dependence.

It is usually beneficial to adaptively set the hyper-parameters during the search procedure in an EA. A great number of EAs with adaptive hyper-parameters have been studied (see~\cite{Aleti2016ystematic} for detail). In most of these studies, it is the numerical hyper-parameters that are made time-variant, e.g. the scaling and crossover rate in DE are adaptively updated during the search by summarizing previous information in JADE~\cite{Zhang2009JADE}. Only recently, some EAs proposed to update structural hyper-parameter adaptively, e.g., the population size in jSO~\cite{Brest2017Single} is designed to be linearly decreasing.




However, there is no principle way to tune structural hyper-parameters. Their tuning is case-sensitive. For instance, controlling the selection of DE operators in SaDE~\cite{Qin2009Differential} can be totally different from the switching between univariate sampling and CMA-ES in HSES. 

Recall that to adaptively configure numerical hyper-parameters, new hyper-parameters are updated based on learning from previous search history as seen in JADE. The updating can be broadly considered as a learning problem. This perspective can be applied for adaptively tuning structural parameters. To implement this idea, we resort to reinforcement learning (RL) by modeling the tuning procedure as a finite-horizon Markov Decision Process~\cite{Puterman1994Markov}.

In this paper, we propose to use RL, specifically the Q-learning algorithm, to tune the structural hyper-parameter of HSES, i.e. the number of iterations used by the first univariate sampling. In the rest of the paper, Section~\ref{preliminaries} briefly review HSES, and concepts and algorithms of RL. The proposed structural hyper-parameter tuning algorithm, called Q-HSES, is presented in Section~\ref{method}. Section~\ref{results} summarizes experimental results on the CEC 2018 test functions in comparison with HSES. The conclusion is given in Section~\ref{conclusion}.



\section{Preliminaries}\label{preliminaries}

\subsection{Reinforcement learning}\label{rl_preliminaries}

\begin{figure}
\centerline{\includegraphics[width=0.95\columnwidth]{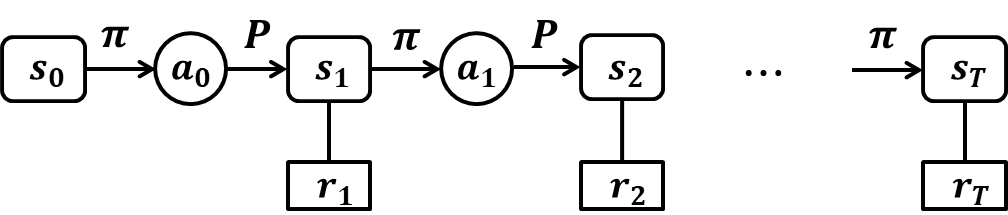}}
\caption{The basic idea and elements involved in a RL model, where $s, a, r$ represents state, action and reward, respectively, and $\pi$ and $P$ represents the policy and state transition probability, respectively. }\label{RL_introduction}
\end{figure}

RL has been playing an important role in the thriving of artificial intelligence. It aims to find a policy so that an agent is able to take optimal actions in an environment. Fig.~\ref{RL_introduction} shows a typical representation of RL model. It can be modeled as a Markov decision process (MDP). Consider a finite-horizon MDP with discrete and finite state and action space defined by the tuple $(\mathcal{S},\mathcal{A},\mu_0,p,r,\pi,T)$ where $\mathcal{S}\in\mathbb{R}^D$ denotes the state space, $\mathcal{A}\in\mathbb{R}^d$ the action space, $\mu_0$ the initial distribution of the state, $r:\mathcal{S}\rightarrow\mathbb{R}$ the reward, and $T$ the time horizon, respectively. At each time $t$, there is an $s_t\in\mathcal{S}$, $a_t\in\mathcal{A}$ and a transition probability $p:\mathcal{S}\times\mathcal{A}\times\mathcal{S}\rightarrow \mathbb{R}$, where $p(s_{t+1}|a_t,s_t)$ denotes the transition probability of $s_{t+1}$ conditionally based on $s_t$ and $a_t$. The policy $\pi:\mathcal{S}\times\mathcal{A}\times \{0,1,\cdots\,T\}\rightarrow\mathbb{R}$, where $\pi(a_t|s_t)$ is the probability of choosing action $a_t$ when observing current state $s_t$. The goal is to find a policy $\pi = p(a_t|s_t)$  so as to maximize the expectation of total rewards $\mathbb{E}\left(\sum_{t=1}^T \alpha_t r_t \right) $ where $\alpha_t$ denote time-step dependent weighting factors. In practice, weighting factors is always set to exponential power of constant i.e. $\alpha_t=\gamma^t$.

There are many RL algorithms, such as Q-learning, sarsa, deep Q network and policy gradient (interested readers please see details in \cite{Sutton1998Reinforcement}), which are developed to deal with different environments. Among them, Q-learning is developed for MDP with discrete state and action space, based on value iteration. Its core idea is to use the action-value function $Q(s,a)$ to estimate the reward in case $s_t = s$ and $a_t = a$, which is defined as
\[Q(s,a) = \mathbb{E}_{\pi}(r_{t+1}+ \max_{a_{t+1}}\gamma Q(s_{t+1}, a_{t+1})|a_t = s, a_t = a)\] As $\mathcal{A}$ is discrete and finite, policy can be regarded as \[\pi(a|s)=\arg\max_{a\in\mathcal{A}}Q(s,a).\] The Q-learning algorithm can be summarized in Alg.~\ref{alg:Q_learning} (taken from~\cite{Sutton1998Reinforcement}). 

\begin{algorithm}[htbp]
\begin{algorithmic}[1]\caption{Q-learning}\label{alg:Q_learning}
\STATE Initialize $Q(s,a) = 0$ for all $a \in {\cal A}, s \in {\cal S}$;\label{q0}
\FOR {$e=1: \text{maxE}$}
    \FOR {$t=0:T$}\label{q1}
        \STATE Choose $a_t$ using policy derived from $Q(s,a)$ ($\varepsilon$-greedy) \label{generate}; \label{q2}
        \STATE Take $a_t$ and observe $s_{t+1}$ and $r_{t+1}$;\label{q3}
        \STATE Compute \begin{multline} Q(s_t,a_t)=(1-\alpha)Q(s_t,a_t)+ \\\alpha [\gamma \max_{a_{t+1}}Q(s_{t+1},a_{t+1})+r_{t+1}];\end{multline}\label{q4}
    \ENDFOR\label{q5}
\ENDFOR
\end{algorithmic}
\end{algorithm}
In Alg.~\ref{alg:Q_learning}, a maximum number ($\text{maxE}$) of epoch is used to learn the action-value function $Q(s,a)$ for each $s \in {\cal S}$ and $a \in {\cal A}$. $Q(s,a)$ is initialized to be zero (line~\ref{q0}). Then at each epoch, a  trajectory is obtained by applying the $\varepsilon$-greedy policy (line~\ref{q1}, which means in probability $1-\varepsilon$, action $a_t $ is taken as $\arg\max Q(s_t, a)$ and in probability $\epsilon$, action $a_t$ is randomly chosen from ${\cal A}$. Based on the state and action at time $t$, the action-value function is updated according to the equation in line~\ref{q4}. The algorithm terminates at the maxE epoch.


\subsection{HSES}\label{HSES_preliminaries}

\begin{figure}
\centerline{\includegraphics[width=1\columnwidth]{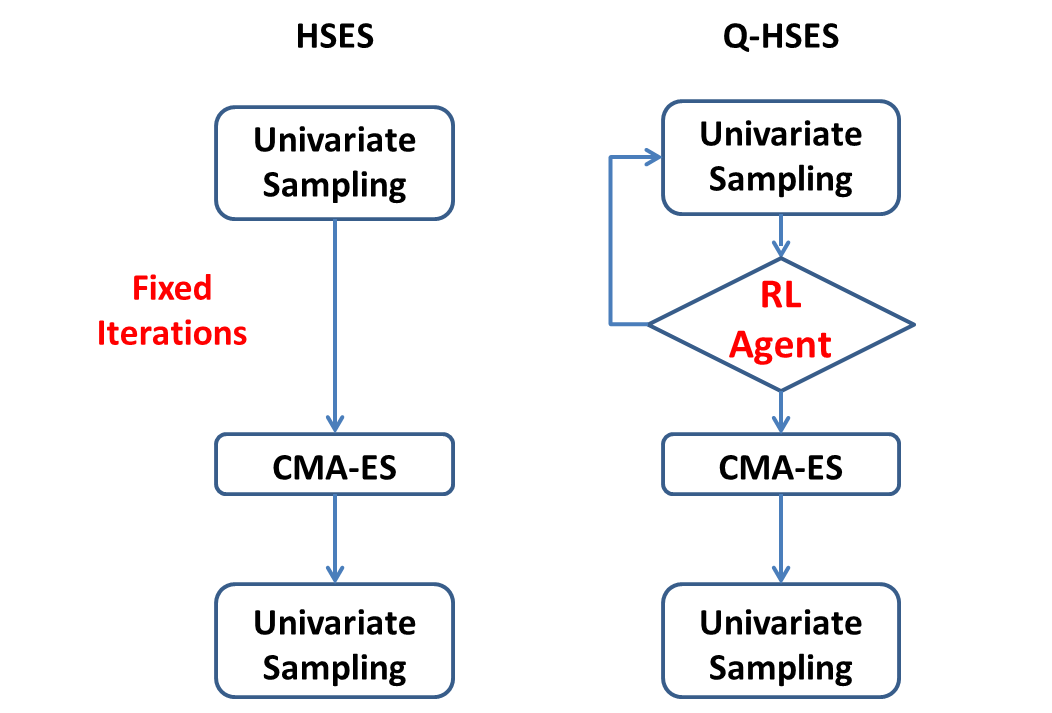}}
\caption{The schematic diagram of HSES and Q-HSES.}\label{QHSES_framework}
\end{figure}

HSES~\cite{Geng2018hygrid} is the winner algorithm in CEC 2018 competition. Its schematic diagram can be seen in the left plot of Fig.~\ref{QHSES_framework}. The pseudo-code of HSES is shown in Alg.~\ref{alg:HSES}.

\begin{algorithm}
\begin{algorithmic}[1]\caption{The pseudo-code of HSES}\label{alg:HSES}
\REQUIRE an optimization function $f(\mathbf{x}), \mathbf{x} \in \mathbb{R}^n$, initial population $\mathbf{X}^0=[\mathbf{x}^0_1,\cdots,\mathbf{x}^0_N]\in \mathbb{R}^{n\times N}$, the switch iteration $\mathbf{I}_1 \in \mathbb{N}_+$, maximum evaluations $\text{MaxNFE}$.
\ENSURE an optimal solution $\mathbf{x}^*$.
\STATE Set $\text{Idx}  \leftarrow \emptyset$;
\STATE $[\mathbf{X}^{t_1}, \text{NFE}_1, \mathbf{x}_1^*]\leftarrow \text{UniSampling}(\mathbf{X}^0, f, \text{Idx}; \mathbf{I}_1)$; \label{a1}
\STATE $[\mathbf{X}^{t_2}, \text{NFE}_2 , \mathbf{x}_2^*] \leftarrow \text{CMA-ES}(\mathbf{X}^{t_1};\theta)$; \label{a2}
\STATE $\text{Idx}  \leftarrow \text{Detect}(\mathbf{X}^{t_2} )$; \label{a3}
\STATE $[\mathbf{X}^{t_3}, \text{NFE}_3 ,\mathbf{x}_3^*]  \leftarrow \text{UniSampling}(\mathbf{X}^{t_2},f, \text{Idx}; \mathbf{I}_2)$; \label{a4}
\RETURN $\mathbf{x}^*  \leftarrow \mathbf{x}_3^{*}$.
\end{algorithmic}
\end{algorithm} In Alg.~\ref{alg:HSES}, the function $\text{UniSampling}(\cdot)$ performs univariate sampling for $\mathbf{I}_1$ iterations and returns a population of solutions $\mathbf{X}^{t_1}$ with relatively high qualities, the number of fitness evaluations used ($\text{NFE}_1 = N \times \mathbf{I}_1$), and the best solution $\mathbf{x}_1^{*}$ found by univariate sampling. $\mathbf{X}^{t_1}$ is then used as the initial population of $\text{CMA-ES}$ with parameter $\theta$. CMA-ES returns a population  $\mathbf{X}^{t_2}$, the fitness evaluations used $\text{NFE}_2$ and the best solution it found $\mathbf{x}_2^*$. Once CMA-ES has terminated, the function $\text{Detect}(\cdot)$ is used to find which variables should be fixed for the next univariate sampling. In the new univariate sampling, a maximum of $\mathbf{I}_2$ iterations is carried out so that the whole number of evaluations is no more than the given constant $\text{MaxNFE}$. Interested readers please refer to~\cite{Geng2018hygrid} for algorithm details. The algorithm returns the best solution found $\mathbf{x}_3^*$ at termination.

In HSES, except the parameters like population size $N$ in univariate sampling and the parameter $\theta$ of CMA-ES, the switch iteration $\mathbf{I}_1$ is vital to algorithmic performance. It determines how much resources are allocated for the first univariate sampling, which reflects the tradeoff between exploration and exploitation.

In our study, we care only on how to determine the structural hyper-parameter $\mathbf{I}_1$. The other hyper-parameters, such as $N$ and $\theta$, are fixed as used in the original paper of HSES.




\section{Method}\label{method}


We model the tuning procedure of structural hyper-parameters as a finite-horizon MDP with discrete state and action space and horizon $T$. At each state, the RL agent chooses an action, i.e. switching to CMA-ES or not. If the action is not to switch, univariate sampling will be carried out. The next state will be observed. Otherwise, if the action is to switch to CMA-ES, the agent will stop and execute the rest of HSES (line~\ref{a2}-\ref{a4} of Alg.~\ref{alg:HSES}). 

In the following, we shall present the components used in RL, including state, action, transition probability and reward. We denote $f_{\text{best}}^k$ the minimum function value obtained up to the $k$-th iteration. 

\textbf{State}: In our RL, $s_t$ includes two parts $s_t^1$ and $s_t^2$ which are defined as follows:
\begin{equation}\label{state}
\begin{split}
s_t^1 &\triangleq  \frac{\log\left(f_{\text{best}}^{10(t-2)}\right)-\log\left(f_{\text{best}}^{10t}\right)}{\left|\log\left(f_{\text{best}}^{10(t-2)}\right)\right|},t>1\\
s_t^2 &\triangleq  \frac{\log\left(f_{\text{best}}^{0}\right)-\log\left(f_{\text{best}}^{10t}\right) }{\left|\log\left(f_{\text{best}}^{0}\right)\right|}
\end{split}
\end{equation}and $s_1^1 \triangleq  \frac{\log\left(f_{\text{best}}^{0}\right)-\log\left(f_{\text{best}}^{10}\right)}{\left|\log\left(f_{\text{best}}^{0}\right)\right|}$. In the definition, $s^1$ is used to measure the difference between the best function values in adjacent 20 steps; $s^2$ measures the descent rate from the first population. We use 10 times $t$ because we don't want to judge switching or not every iteration but every 10 iterations.

It is seen that the range of $s^1$ and $s^2$ are all in $[0, +\infty)$. To make the state space discrete and finite, we divide the range to $[0,0.005]$,$(0.005,0.05]$,$(0.5,0.09]$,$(0.09,0.5]$,$(0.5,1]$,$(1,+\infty)$ for $s^1$, and $[0,0.2]$,$(0.2,0.5]$,$(0.5,0.8]$,$(0.8,1.2]$,$(1.2,3]$,$(3,+\infty)$ for $s^2$.

\textbf{Action:} The action space is $\{0,1\}$. That is, the agent can either choose to switch (action equals to 1) to CMA-ES or do not switch (action equals to 0).

\textbf{Transition probability}: When $t<T$ and $a_t$ is 0, the next state $s_{t+1}$ is defined as mentioned above. When $t\geq T$ or $a_t$ is 1, $s_{t+1}$ will be the ``terminal state". Here $T$ is the horizon. It constrains the maximum iterations that can be used by univariate sampling. In this paper, we set $T=20$.

\textbf{Reward}: In the terminal state, the reward will be the logarithm of the minimum function value found by the search. Otherwise, the reward is set to zero since the algorithm's performance is not known before the terminal state. 

\subsection{The proposed Q-learning based HSES}

Given above definitions, the Q-learning algorithm, i.e. Alg.~\ref{alg:Q_learning}, can be used to train the agent. However, directly using Q-learning can result in the following three problems. First, to make Q-learning converge, the maximal number of epoch, $\text{MaxE}$, is generally large. This means HSES needs to be executed many times, which is not acceptable. Second, in Q-learning, the action value function is updated at each trajectory based on current policy. Since EAs are stochastic, the performance of one trajectory is not stable, which means the learning of the action value function is not efficient. Third, the trajectory created based on current policy is unbalanced (as using $\varepsilon$-greedy policy). That is, some states may be rarely observed in the trajectories. This can make the learning converge slowly.
\begin{algorithm}[htbp]
\begin{algorithmic}[1]\caption{The training process for the RL agent.}\label{alg:agent}
\REQUIRE Training functions $f_1,\cdots,f_L$, the maximal epoch $\text{maxE}$, the horizon $T$ and the learning rate $\alpha$.
\ENSURE an optimal policy $\pi(a|s)$.
\STATE Initialize $\text{meta}_Q(s,a) = 0$ for all $s \in {\cal S}, a\in \{0,1\}$;\label{t1}
\FOR {$l=1:L$}
\STATE Initialize $Q(s,a) = 0$ for all $s \in {\cal S}, a\in \{0,1\}$;
\STATE Create $T$ training trajectories $\text{Tr}_m,\ m=1,\cdots, T$;\label{Generate_data}
    \FOR {$e=1:\text{maxE}$}
        \FOR{each trajectory $m$}
            \FOR {$t=1:T$}
                \STATE $Q(s_t,a_t)\leftarrow(1-\alpha)Q(s_t,a_t)+ \alpha\gamma\max_{a_{t+1}}Q(s_{t+1},a_{t+1})+r_{t+1}$ where $\{s_t,a_t\}\in \text{Tr}_m$;\label{QQ_update}
            \ENDFOR
        \ENDFOR
    \ENDFOR
    \FOR {$s\in \cal{S}$} \label{scale_Q_start}
        \IF {$Q(s,1)>Q(s,0)$}
            \STATE$\text{meta}_Q(s,1) \leftarrow \text{meta}_Q(s,1)+1$;
        \ENDIF
        \IF{$Q(s,1)<Q(s,0)$}
            \STATE $\text{meta}_Q(s,0) \leftarrow \text{meta}_Q(s,0)+1$;
        \ENDIF
    \ENDFOR \label{scale_Q_end}
\ENDFOR
\RETURN $\pi(a|s)\leftarrow\arg\max_a \text{meta}_Q(s,a)$.\label{tend}
\end{algorithmic}
\end{algorithm}

Our training process is summarized in Alg.~\ref{alg:agent}. In line~\ref{Generate_data}, we generate $T$ trajectories $\text{Tr}_m,\ m=1,2,\cdots,T$. Each trajectory corresponds to the average of 51 runs of the HSES algorithm with switch iteration $\mathbf{I}_1$ to be $10,20,\cdots,200$ for a given training function. With these trajectories, the action value function $Q(s,a)$ is updated as seen in line~\ref{QQ_update}. To eliminate the influence of different function scales, an auxiliary action-value function $\text{meta}_Q(s,a)$ is applied. It records the rank relationship between $Q(s,0)$ and $Q(s,1)$ from line~\ref{scale_Q_start} to line~\ref{scale_Q_end} for each training function. After training, the policy $\pi(a|s)$ is set to be $\arg\max_a \text{meta}_Q(s,a)$ (line~\ref{tend}). Note that for a state $s$, $\text{meta}_Q(s,0)=\text{meta}_Q(s,1)$ means that there is no evidence to tell which action (switch or not) is better. If this is the case, an action will be randomly chosen.

The reason that the proposed training process can handle the mentioned learning problems is simply because the created trajectories $\text{Tr}_m$ contains all possible situations for the switch iteration. That is, the performance of HSES with every possible switch iteration from 10 to 200 is observed, which is measured by the average of 51 HSES runs. This can make the training steady and the observed states balanced. Since we do not need to sample new trajectories during training, the training efficiency can be guaranteed.

Once the agent is learned (i.e. the optimal $\pi$ is found), it is embedded within the HSES algorithm. The resultant algorithm is called Q-HSES. The right plot of Fig.~\ref{QHSES_framework} shows the digram of Q-HSES. Alg.~\ref{alg:Q_HSES} summarizes the algorithm. In the algorithm, after univariate sampling runs for $M$ iterations (line~\ref{qhses1}), the current state $s_t$ is computed (line~\ref{qhses2}) and an action is taken (line~\ref{qhses3}). If action is 0, it returns to the univariate sampling. The procedure repeats until action is 1, which means that the algorithm switches to CMA-ES. The rest operations are the same as in Alg.~\ref{alg:HSES}.
\begin{algorithm}
\begin{algorithmic}[1]\caption{The pseudo-code of Q-HSES}\label{alg:Q_HSES}
\REQUIRE an optimization function $f(\mathbf{x}), \mathbf{x} \in \mathbb{R}^n$, initial population $\mathbf{X}^0=[\mathbf{x}^0_1,\cdots,\mathbf{x}^0_N]\in \mathbb{R}^{n\times N}$, maximum evaluations $\text{MaxNFE}$ and the learned policy $\pi(a|s)$.
\ENSURE an optimal solution $\mathbf{x}^*$.
\STATE Set $\text{Idx} \leftarrow \emptyset, t  \leftarrow 0, \text{NFE}_1  \leftarrow 0$;
\REPEAT
\STATE $[\mathbf{X}^{t+1}, \text{NFE}_u, \mathbf{x}_1^{*}]\leftarrow \text{UniSampling}(\mathbf{X}^t, f, \text{Idx}; M);$\; \label{qhses1}
\STATE Compute the state $s_t $ by (\ref{state}); \label{qhses2}
\STATE Take $a_t \sim \pi (\cdot|s_t)$; \label{qhses3}
\IF{$a_t = 1 $} \label{qhses4}
\STATE Exit;
\ELSE
\STATE $t \leftarrow t+1$, $\text{NFE}_1 \leftarrow \text{NFE}_1 +  \text{NFE}_u $;
\ENDIF \label{qhses5}
\UNTIL{$t \geq T $}
\STATE $[\mathbf{X}^{t_2}, \text{NFE}_2, \mathbf{x}_2^{*} ] \leftarrow \text{CMA-ES}(\mathbf{X}^{t+1};\theta)$;
\STATE $\text{Idx} = \text{Detect}(\mathbf{X}^{t_2} )$; \label{a3}
\STATE $[\mathbf{X}^{t_3}, \text{NFE}_3,\mathbf{x}_3^{*}]  \leftarrow \text{UniSampling}(\mathbf{X}^{t_2},f, \text{Idx}; \mathbf{I}_2)$;
\RETURN $\mathbf{x}^* \leftarrow \mathbf{x}_3^*$.
\end{algorithmic}
\end{algorithm}

\section{Experiments}\label{results}

In this paper, the 29 functions in the CEC 2018 competition ($f_1-f_{30}$ except $f_2$) is used as benchmark. Metrics, including the rank and average function value, used in the competition are adopted for comparison. The proposed algorithm, Q-HSES, is compared with HSES.

\subsection{Training}

\textbf{Training:} The CEC 2018 test functions contain unimodal functions $f_1$ and $f_3$, multimodal functions $f_4-f_{10}$, hybrid functions $f_{11}-f_{20}$ and composition functions $f_{21}-f_{30}$. Our training functions cover these 4 function types. 11 CEC 2018 functions ($f_1,f_6,f_7,f_8, f_{10},f_{14},f_{15},f_{17},f_{18},f_{20},f_{24}$) are used as the training functions. Parameters used in the Q-learning are set as $\gamma=1$, $\alpha=10^{-4}$, $M = 10$, $T = 20$ and $\text{MaxE} =100,000$. As 50D and 100D problems are more complicated and the performance of algorithm is worse than 10D and 30D, the division to the range of $s^2$ is a little bit forward. For 50D and 100D, the interval node of $s^2$ is multiplied by 0.05 and 0.025. Except the switch iteration, the other hyper-parameters are held the same as HSES (such as population size and the parameters of CMA-ES).

\textbf{Testing:} Q-HSES is used to optimize all the 29 functions of CEC 2018.

\subsection{Comparison Results}

In the original HSES, the switch iteration is fixed to be 100. The statistics of the obtained results of Q-HSES on test functions of CEC 2018 are summarized in Tables~\ref{10D},\ref{30D},\ref{50D} and \ref{100D} for 10D, 30D, 50D and 100D, respectively, in which the best, worst, mean and std. values are given.

Each of the statistics is obtained over 200, 100, 51, 51 runs on the error values (i.e. the difference between the obtained optimum and the known global optimum). When the error values are smaller or equal to $10^{-8}$, they are treated as 0.

\begin{table}[htbp]
\caption{Results in 10D for 200 runs obtained by Q-HSES.}\label{10D}
\centering
\begin{tabular}{|c|c|c|c|c|c|}
\hline
 &best & worst & median &mean &std\\
 \hline
 $f_1$ &0.00$e$+00 & 0.00$e$+00 & 0.00$e$+00 &0.00$e$+00&0.00$e$+00\\
 \hline
  $f_3$ &0.00$e$+00 & 0.00$e$+00 & 0.00$e$+00 &0.00$e$+00&0.00$e$+00\\
 \hline
  $f_4$ &0.00$e$+00 & 0.00$e$+00 & 0.00$e$+00 &0.00$e$+00&0.00$e$+00\\
 \hline
  $f_5$ &0.00$e$+00 & 3.98$e$+00 & 9.95$e$-01 &8.40$e$-01&8.05$e$-01\\
 \hline
  $f_6$ &0.00$e$+00 & 0.00$e$+00 & 0.00$e$+00 &0.00$e$+00&0.00$e$+00\\
 \hline
  $f_7$ &1.04$e$+01 & 1.28$e$+01 & 1.10$e$+01 &1.11$e$+01&4.96$e$-01\\
 \hline
  $f_8$ &0.00$e$+00 & 2.98$e$+00 & 0.00$e$+00 &5.87$e$-01&7.54$e$-01\\
 \hline
  $f_9$ &0.00$e$+00 & 0.00$e$+00 & 0.00$e$+00 &0.00$e$+00&0.00$e$+00\\
 \hline
  $f_{10}$ &6.25$e$-02 & 6.12$e$+02 & 3.73$e$+00 &1.07$e$+02&1.42$e$+02\\
 \hline
  $f_{11}$ &0.00$e$+00 & 1.99$e$+00 & 0.00$e$+00 &1.29$e$-01&3.77$e$-01\\
 \hline
  $f_{12}$ &0.00$e$+00 & 4.14$e$+02 & 2.09$e$-01 &2.43$e$+01&6.42$e$+01\\
 \hline
  $f_{13}$ &0.00$e$+00 & 8.51$e$+00 & 5.20$e$+00 &3.39$e$+00&2.55$e$+00\\
 \hline
  $f_{14}$ &0.00$e$+00 & 3.11$e$+02 & 9.85$e$-04 &7.28$e$+00&3.25$e$+01\\
 \hline
  $f_{15}$ &6.71$e$-05 & 5.10$e$+00 & 4.35$e$-01 &5.58$e$-01&7.08$e$-01\\
 \hline
  $f_{16}$ &1.17$e$-01  &1.19$e$+02	& 7.28$e$-01 &1.94$e$+00&1.18$e$+01\\
 \hline
  $f_{17}$ &1.97$e$-02	&3.87$e$+01	&1.73$e$+01	&1.27$e$+01	&1.07$e$+01\\
  \hline
 $f_{18}$ &2.24$e$-05	&2.15$e$+01	&9.39$e$-01	&2.15$e$+00	&3.25$e$+00\\
  \hline
 $f_{19}$&2.44$e$-02	&8.99$e$+00	&1.74$e$-01	&5.41$e$-01	&1.34$e$+00\\
  \hline
 $f_{20}$&0.00$e$+00	&1.20$e$+02	&1.31$e$+00	&8.82$e$+00	&1.81$e$+01\\
  \hline
 $f_{21}$&1.00$e$+02	&2.06$e$+02	&2.02$e$+02	&1.97$e$+02	&2.13$e$+01\\
  \hline
 $f_{22}$&1.00$e$+02	&1.00$e$+02	&1.00$e$+02	&1.00$e$+02	&2.85$e$-14\\
 \hline
 $f_{23}$&3.00$e$+02	&3.06$e$+02	&3.00$e$+02	&3.01$e$+02	&1.51$e$+00\\
 \hline
 $f_{24}$&1.00$e$+02	&3.32$e$+02	&3.29$e$+02	&3.27$e$+02	&1.62$e$+01\\
 \hline
 $f_{25}$&3.98$e$+02	&4.50$e$+02	&4.46$e$+02	&4.46$e$+02	&3.62$e$+00\\
 \hline
 $f_{26}$&2.00$e$+02	&3.00$e$+02	&3.00$e$+02	&3.00$e$+02	&7.07$e$+00\\
 \hline
 $f_{27}$&3.91$e$+02	&4.01$e$+02	&3.98$e$+02	&3.97$e$+02	&1.95$e$+00\\
 \hline
 $f_{28}$&3.00$e$+02	&6.46$e$+02	&5.84$e$+02	&5.92$e$+02	&4.77$e$+01\\
 \hline
 $f_{29}$&2.41$e$+02	&2.97$e$+02	&2.63$e$+02	&2.64$e$+02	&1.00$e$+01\\
 \hline
 $f_{30}$&3.95$e$+02	&4.84$e$+02	&3.95$e$+02	&4.11$e$+02	&2.21$e$+01\\
 \hline
\end{tabular}
\end{table}

\begin{table}[htbp]
\caption{Results in 30D for 100 runs obtained by Q-HSES.}\label{30D}
\centering
\begin{tabular}{|c|c|c|c|c|c|}
\hline
 &best & worst & median &mean &std\\
 \hline
 $f_1$ &0.00$e$+00 & 0.00$e$+00 & 0.00$e$+00 &0.00$e$+00&0.00$e$+00\\
 \hline
  $f_3$ &0.00$e$+00	&0.00$e$+00	&0.00$e$+00	&0.00$e$+00	&0.00$e$+00\\
 \hline
  $f_4$ &0.00$e$+00	&3.99$e$+00	&3.99$e$+00	&2.91$e$+00	&1.78$e$+00\\
 \hline
  $f_5$ &2.98$e$+00	&1.19$e$+01	&6.96$e$+00	&6.77$e$+00	&2.04$e$+00\\
 \hline
  $f_6$ &0.00$e$+00	&0.00$e$+00	&0.00$e$+00	&0.00$e$+00	&0.00$e$+00\\
 \hline
  $f_7$ &3.29$e$+01	&4.04$e$+01	&3.50$e$+01	&3.53$e$+01	&1.43$e$+00\\
 \hline
  $f_8$ &2.98$e$+00	&1.09$e$+01	&6.96$e$+00	&6.64$e$+00	&1.92$e$+00\\
 \hline
  $f_9$ &0.00$e$+00	&0.00$e$+00 &0.00$e$+00	&0.00$e$+00	&0.00$e$+00\\
 \hline
  $f_{10}$ &1.27$e$+02	&1.81$e$+03	&8.53$e$+02	&8.69$e$+02	&3.44$e$+02\\
 \hline
  $f_{11}$ &0.00$e$+00	&7.29$e$+01	&4.97$e$+00	&1.22$e$+01	&2.01$e$+01\\
 \hline
  $f_{12}$ &6.96$e$-02	&3.59$e$+02	&3.93$e$+00	&2.89$e$+01	&7.39$e$+01\\
 \hline
  $f_{13}$ &3.81$e$+00	&8.59$e$+01	&2.65$e$+01	&3.08$e$+01	&1.57$e$+01\\
 \hline
  $f_{14}$ &1.47$e$-04	&3.58$e$+01	&1.40$e$+01	&1.18$e$+01	&9.89$e$+00\\
 \hline
  $f_{15}$ &3.92$e$-01	&3.45$e$+01	&3.49$e$+00	&4.45$e$+00	&4.59$e$+00\\
 \hline
  $f_{16}$ &1.26$e$+00	&8.72$e$+02	&2.43$e$+02	&2.64$e$+02	&2.00$e$+02\\
 \hline
  $f_{17}$ &2.37$e$+00	&6.03$e$+02	&2.48$e$+01	&7.15$e$+01	&1.12$e$+02\\
 \hline
  $f_{18}$ &4.98$e$-01	&2.74$e$+01	&2.07$e$+01	&1.89$e$+01	&6.15$e$+00\\
 \hline
  $f_{19}$ &1.72$e$+00	&3.10$e$+01	&3.49$e$+00	&4.25$e$+00	&4.20$e$+00\\
 \hline
  $f_{20}$ &1.20$e$+02	&4.21$e$+02	&1.43$e$+02	&1.62$e$+02	&6.01$e$+01\\
 \hline
  $f_{21}$ &2.02$e$+02	&2.21$e$+02	&2.07$e$+02	&2.08$e$+02	&3.41$e$+00\\
 \hline
  $f_{22}$ &1.00$e$+02	&1.00$e$+02	&1.00$e$+02	&1.00$e$+02	&3.03$e$-13\\
 \hline
  $f_{23}$ &3.37$e$+02	&3.69$e$+02	&3.50$e$+02	&3.51$e$+02	&8.18$e$+00\\
 \hline
  $f_{24}$ &4.08$e$+02	&4.32$e$+02	&4.20$e$+02	&4.20$e$+02	&4.70$e$+00\\
 \hline
  $f_{25}$ &3.87$e$+02	&3.87$e$+02	&3.87$e$+02	&3.87$e$+02	&2.38$e$-02\\
 \hline
  $f_{26}$ &2.00$e$+02	&1.47$e$+03	&9.17$e$+02	&8.84$e$+02	&2.04$e$+02\\
 \hline
  $f_{27}$ &5.07$e$+02	&5.50$e$+02	&5.24$e$+02	&5.25$e$+02	&1.02$e$+01\\
 \hline
  $f_{28}$ &3.00$e$+02	&4.03$e$+02	&3.00$e$+02	&3.19$e$+02	&3.99$e$+01\\
 \hline
  $f_{29}$ &4.09$e$+02	&8.11$e$+02	&4.39$e$+02	&4.68$e$+02	&7.66$e$+01\\
 \hline
  $f_{30}$ &1.97$e$+03	&2.20$e$+03	&2.06$e$+03	&2.06$e$+03	&4.45$e$+01\\
 \hline
\end{tabular}
\end{table}

\begin{table}[htbp]
\caption{Results in 50D for 51 runs obtained by Q-HSES.}\label{50D}
\centering
\begin{tabular}{|c|c|c|c|c|c|}
\hline
 &best & worst & median &mean &std\\
 \hline
 $f_1$ &0.00$e$+00 & 1.70$e$-08 & 0.00$e$+00 &0.00$e$+00&0.00$e$+00\\
 \hline
  $f_3$ &0.00$e$+00 & 0.00$e$+00 & 0.00$e$+00 &0.00$e$+00&0.00$e$+00\\
 \hline
  $f_4$ &0.00$e$+00 & 1.14$e$+02	&2.85$e$+01	&4.56$e$+01	&4.72$e$+01 \\
 \hline
  $f_5$ &0.00$e$+00 & 3.98$e$+00	&9.95$e$-01	&1.11$e$+00	&1.08$e$+00\\
 \hline
  $f_6$ &3.59$e$-08	&2.95$e$-05	&1.05$e$-05	&1.27$e$-05	&1.08$e$-05\\
 \hline
  $f_7$ &5.40$e$+01	&5.81$e$+01	&5.53$e$+01	&5.54$e$+01	&8.05$e$-01\\
 \hline
  $f_8$ &0.00$e$+00	&3.98$e$+00	&9.95$e$-01	&1.44$e$+00	&9.81$e$-01\\
 \hline
  $f_9$ &0.00$e$+00 & 0.00$e$+00 & 0.00$e$+00 &0.00$e$+00&0.00$e$+00\\
 \hline
  $f_{10}$ &1.24$e$+02	&7.39$e$+02	&1.33$e$+02	&2.60$e$+02	&1.69$e$+02\\
 \hline
  $f_{11}$ &1.83$e$+01	&2.62$e$+01	&2.33$e$+01	&2.30$e$+01	&1.90$e$+00\\
 \hline
  $f_{12}$ &2.40$e$+00	&4.08$e$+02	&1.33$e$+02	&1.53$e$+02	&1.23$e$+02\\
 \hline
  $f_{13}$ &2.54$e$-06	&7.85$e$+01	&4.54$e$+01	&3.93$e$+01	&2.54$e$+01\\
 \hline
  $f_{14}$ &1.24$e$-04	&2.24$e$+01	&2.03$e$+01	&1.41$e$+01	&9.51$e$+00\\
 \hline
  $f_{15}$ &3.14$e$+00	&1.85$e$+01	&1.74$e$+01	&1.73$e$+01	&2.05$e$+00\\
 \hline
  $f_{16}$ &1.28$e$+02	&1.24$e$+03	&7.62$e$+02	&6.92$e$+02	&2.73$e$+02\\
 \hline
  $f_{17}$ &2.96$e$+01	&9.75$e$+02	&1.77$e$+02	&2.74$e$+02	&1.70$e$+02\\
 \hline
  $f_{18}$ &5.72$e$-01	&2.11$e$+01	&2.09$e$+01	&2.05$e$+01	&2.85$e$+00\\
 \hline
  $f_{19}$ &3.57$e$+00  &2.61$e$+01 &5.55$e$+00 &7.13$e$+00 &4.79$e$+00\\
 \hline
  $f_{20}$ &2.04$e$+01	&2.44$e$+02	&2.50$e$+01	&3.45$e$+01	&4.25$e$+01\\
 \hline
  $f_{21}$ &2.01$e$+02	&2.10$e$+02	&2.05$e$+02	&2.05$e$+02	&1.21$e$+00\\
 \hline
  $f_{22}$ &1.00$e$+02	&1.00$e$+02	&1.00$e$+02	&1.00$e$+02	&1.16$e$-06\\
 \hline
  $f_{23}$ &4.04$e$+02	&4.41$e$+02	&4.24$e$+02	&4.25$e$+02	&9.29$e$+00\\
 \hline
  $f_{24}$ &4.84$e$+02	&4.94$e$+02	&4.90$e$+02	&4.89$e$+02	&2.51$e$+00\\
 \hline
  $f_{25}$ &4.71$e$+02	&5.66$e$+02	&5.63$e$+02	&5.49$e$+02	&2.38$e$+01\\
 \hline
  $f_{26}$ &4.00$e$+02	&8.75$e$+02	&6.58$e$+02	&6.16$e$+02	&1.45$e$+02\\
 \hline
  $f_{27}$ &5.24$e$+02	&6.07$e$+02	&5.52$e$+02	&5.56$e$+02	&2.18$e$+01\\
 \hline
  $f_{28}$ &4.70$e$+02	&5.08$e$+02	&5.08$e$+02	&5.00$e$+02	&1.20$e$+01\\
 \hline
  $f_{29}$ &3.03$e$+02	&8.19$e$+02	&3.47$e$+02	&4.50$e$+02	&1.50$e$+02\\
 \hline
  $f_{30}$ &5.80$e$+05	&6.71$e$+05	&5.96$e$+05	&5.99$e$+05	&1.56$e$+04\\
 \hline
\end{tabular}

\end{table}
\begin{table}[htbp]
\caption{Results in 100D for 51 runs obtained by Q-HSES.}\label{100D}
\centering
\begin{tabular}{|c|c|c|c|c|c|}
\hline
 &best & worst & median &mean &std\\
 \hline
 $f_1$ &0.00$e$+00 & 2.45$e$-08 & 0.00$e$+00 &0.00$e$+00&0.00$e$+00\\
 \hline
  $f_3$ &0.00$e$+00 & 3.88$e$-08 & 0.00$e$+00 &0.00$e$+00&0.00$e$+00\\
 \hline
  $f_4$ &0.00$e$+00 & 6.88$e$+01	&0.00$e$+00	&7.74$e$+00	&1.92$e$+01 \\
 \hline
  $f_5$ &9.95$e$-01	&6.96$e$+00	&2.98$e$+00	&3.58$e$+00	&1.49$e$+00\\
 \hline
  $f_6$ &6.51$e$-08	&1.19$e$-07	&8.95$e$-08	&9.23$e$-08	&1.76$e$-08\\
 \hline
  $f_7$ &1.08$e$+02	&1.13$e$+02	&1.10$e$+02	&1.10$e$+02	&1.19$e$+00\\
 \hline
  $f_8$ &9.95$e$-01	&7.96$e$+00	&3.98$e$+00	&3.98$e$+00	&2.14$e$+00\\
 \hline
  $f_9$ &0.00$e$+00	&6.22$e$+00	&0.00$e$+00	&7.88$e$-01	&1.61$e$+00\\
 \hline
  $f_{10}$ &7.63$e$+02	&2.04$e$+03	&1.23$e$+03	&1.31$e$+03	&3.07$e$+02\\
 \hline
  $f_{11}$ &1.69$e$-06	&8.91$e$+01	&1.27$e$+01	&3.55$e$+01	&3.80$e$+01\\
 \hline
  $f_{12}$ &3.43$e$+02	&1.97$e$+03	&8.70$e$+02	&9.13$e$+02	&4.13$e$+02\\
 \hline
  $f_{13}$ &3.77$e$+01	&5.96$e$+01	&4.15$e$+01	&4.32$e$+01	&5.65$e$+00\\
 \hline
  $f_{14}$ &1.99$e$+00	&2.38$e$+01	&2.18$e$+01	&2.11$e$+01	&4.54$e$+00\\
 \hline
  $f_{15}$ &6.89$e$+01	&1.33$e$+02	&9.06$e$+01	&9.87$e$+01	&1.96$e$+01\\
 \hline
  $f_{16}$ &4.31$e$+02	&1.72$e$+03	&1.10$e$+03	&1.04$e$+03	&4.04$e$+02\\
 \hline
  $f_{17}$ &5.80$e$+01	&8.72$e$+02	&4.84$e$+02	&4.96$e$+02	&2.37$e$+02\\
 \hline
  $f_{18}$ &5.19$e$-01	&2.26$e$+01	&1.53$e$+00	&9.41$e$+00	&9.87$e$+00\\
 \hline
  $f_{19}$ &1.10$e$+01	&3.16$e$+01	&1.40$e$+01	&1.50$e$+01	&5.00$e$+00\\
 \hline
  $f_{20}$ &2.77$e$+02	&1.33$e$+03	&5.09$e$+02	&5.52$e$+02	&2.78$e$+02\\
 \hline
  $f_{21}$ &2.17$e$+02	&2.32$e$+02	&2.27$e$+02	&2.26$e$+02	&4.01$e$+00\\
 \hline
  $f_{22}$ &1.00$e$+02	&1.00$e$+02	&1.00$e$+02	&1.00$e$+02	&2.40$e$-09\\
 \hline
  $f_{23}$ &5.28$e$+02	&5.58$e$+02	&5.43$e$+02	&5.45$e$+02	&7.51$e$+00\\
 \hline
  $f_{24}$ &8.32$e$+02	&8.52$e$+02	&8.45$e$+02	&8.43$e$+02	&5.97$e$+00\\
 \hline
  $f_{25}$ &6.58$e$+02	&7.87$e$+02	&7.41$e$+02	&7.34$e$+02	&3.36$e$+01\\
 \hline
  $f_{26}$ &2.15$e$+03	&2.55$e$+03	&2.37$e$+03	&2.37$e$+03	&1.23$e$+02\\
 \hline
  $f_{27}$ &6.27$e$+02	&6.47$e$+02	&6.37$e$+02	&6.37$e$+02	&6.01$e$+00\\
 \hline
  $f_{28}$ &3.00$e$+02	&6.33$e$+02	&5.35$e$+02	&5.03$e$+02	&1.10$e$+02\\
 \hline
  $f_{29}$ &8.36$e$+02	&2.14$e$+03	&1.32$e$+03	&1.31$e$+03	&3.71$e$+02\\
 \hline
  $f_{30}$ &2.51$e$+03	&2.91$e$+03	&2.71$e$+03	&2.72$e$+03	&1.16$e$+02\\
 \hline
\end{tabular}
\end{table}


The rank sum hypothesis test is carried out at 5\% significance level between the results obtained by Q-HSES and HSES. The results are summarized in Table~\ref{comparison_rank}. From Table~\ref{comparison_rank}, we see that Q-HSES performs significantly better than HSES in general for all dimensions.

Table~\ref{comparison_funcva} shows the average function values obtained for the test functions in four different dimensions. It is clear that the average function value obtained by Q-HSES is smaller than HSES in 10D, 30D, 50D, but a little bit greater in 100D. 

Overall, we may conclude that the proposed control algorithm based on Q-learning can indeed find a better structural hyper-parameter setting for HSES. 

\begin{table}[h]
\caption{Rank comparison of Q-HSES and HSES}\label{comparison_rank}
\centering
\begin{tabular}{|c|c|c|c|}
\hline
 &better & $\approx$ & worse\\
 \hline
10D &8 & 20 & 1 \\
\hline
30D &7 & 20 & 2 \\
\hline
50D &3 & 24 & 2 \\
\hline
100D &3 & 25 & 1 \\
\hline
\end{tabular}
\end{table}

\begin{table}[h]
\caption{Function value comparison of Q-HSES and HSES}\label{comparison_funcva}
\centering
\begin{tabular}{|c|c|c|}
\hline
 &Q-HSES & HSES\\
 \hline
10D &3504.1& 3511.1  \\
\hline
30D &7246.6 & 7364.5  \\
\hline
50D &604390.5 & 611750.5 \\
\hline
100D &14648.3 & 14602.3 \\
\hline
\end{tabular}
\end{table}


\subsection{Validation}

The switch iteration is fixed in HSES as 100. To validate the performance of the learned agent, Q-HSES is compared with HSES on different switch iterations in 10D. Table~\ref{ablation_study} summarizes the results, in which the hypothesis test results at 5\% significant level are listed. The indices of the functions that Q-HSES performs better are also shown. The indices of the functions that belong to the training functions are typeset in bold.

From Table~\ref{ablation_study}, it is seen that with different switch iterations, Q-HSES performs better than HSES in general: Q-HSES performs better on more functions than HSES. This indicates that the learned agent works well. It is seen that among the functions that Q-HSES performs better, half of them do not belong to the training functions. This indicates the learned agent generalizes well.

\begin{table}[h]
\caption{Validation of Q-HSES }\label{ablation_study}
\centering
\begin{tabular}{|c|c|c|}
\hline
fixed iterations & better / $\approx$ /  worse	&  index\\
 \hline
30 &12/ 14 / 3 & 5 \textbf{7} \textbf{8} \textbf{10} 11 \textbf{17} \textbf{18} \textbf{20}	21	22	23	26  \\
\hline
50 &10/ 17 / 2 & 5	 \textbf{7} \textbf{8} \textbf{10}	11	16	\textbf{17}	\textbf{20}	23	26 \\
\hline
120 & 7/ 20 /  2 &  \textbf{7}	16	\textbf{17}	19	\textbf{20}	23	26\\
\hline
160 & 7/ 20 /  2& 13	\textbf{14}	\textbf{18}	19	22	23	26 \\
\hline
200 & 5/ 24 /  1 & 13	 \textbf{14}	 \textbf{15}	 \textbf{18}	 19 \\
\hline

\end{tabular}
\end{table}

\subsection{Training Process}

\begin{figure*}[htbp]
\centerline{\includegraphics[width=2\columnwidth]{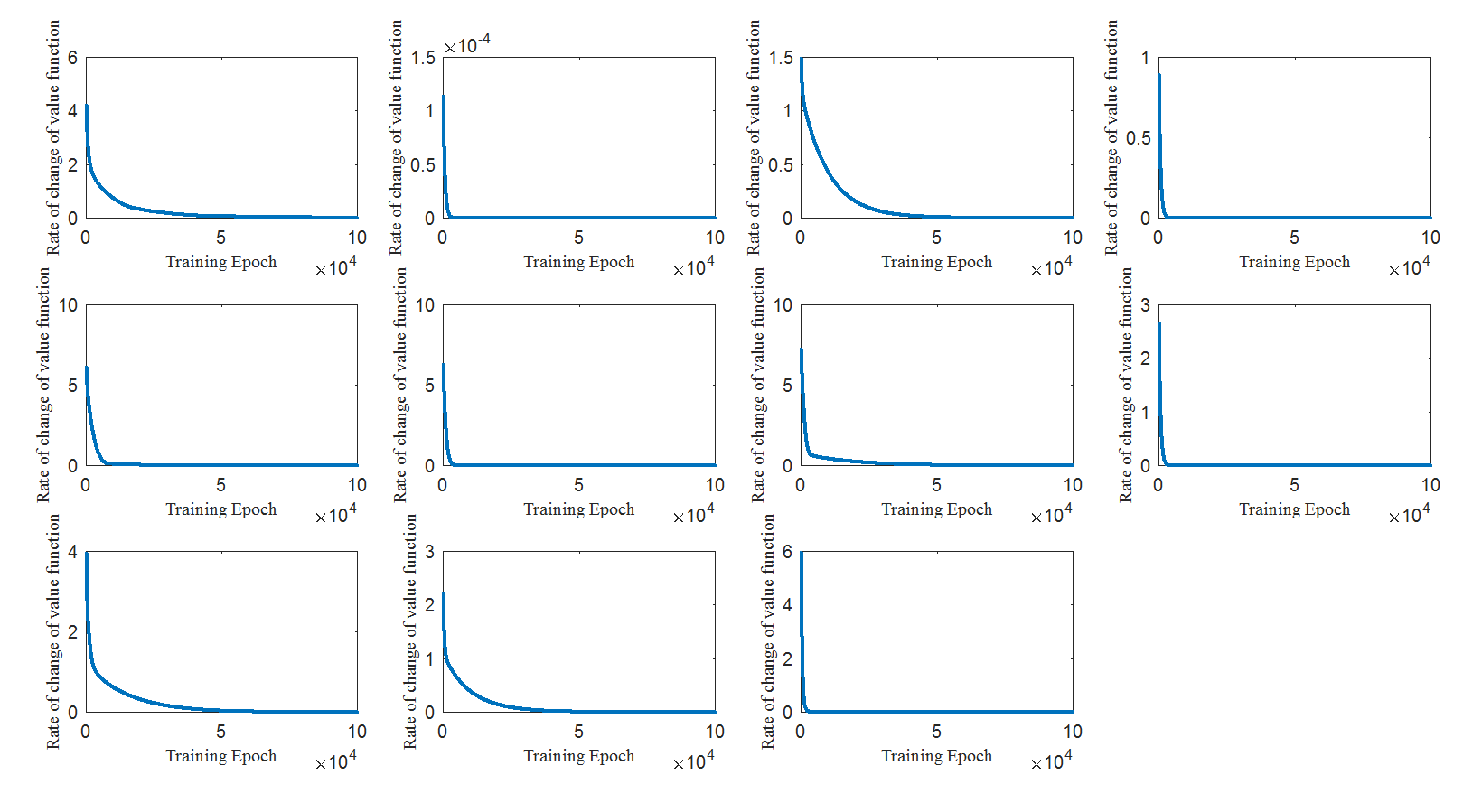}}\label{convergence}
\caption{The rate of function value change against epoch for the 11 training functions.}
\end{figure*}

To verify the training process is convergent, we define a criterion named ``the rate of action value change". As the state space is divided into 36 intervals (6 intervals for $s^1$ and 6 intervals for $s^2$) and action is $\{0,1\}$, then the action-value function $Q(s,a)$ is a $36\times 2$ matrix. Regarding it as a 72D vector, the rate of action value change is defined as the Frobenius norm between two adjacent epoch:
\begin{align*}
\|Q^e(s,a)-Q^{e+1}(s,a)\|_{F}^2
\end{align*}where $Q^e(s,a)$ and $Q^{e+1}(s,a)$ is the action value function at the $e$th and $(e+1)$th epoch, respectively. This definition can be used to show how the training goes. Fig.~\ref{convergence} shows the rates of change for the training functions. From the figure, it is clear that the training process of action value function is convergent. It also shows the range of the changes is very much different  among different functions which roots from different function scales.

\section{Conclusion}\label{conclusion}

In this paper, we first categorized the hyper-parameters of EAs from two perspectives. We then proposed to model the adaptive control of the structural hyper-parameters as a Markov decision process. Based on the formalization, Q-learning was applied to learn an agent for time-variant hyper-parameter tuning for the winner algorithm of CEC 2018, called HSES. We proposed the fundamental elements of the Q-learning for the agent, including states, action and reward. In the experiments, we trained the agent on a selection of functions from the CEC 2018 competition. By embedding the learn agent, Q-HSES was developed. The comparison between Q-HSES and HSES showed that the structural hyper-parameter in HSES controlled by the learned agent performs generally better than HSES.

As a first attempt to use Q-learning for hyper-parameter controlling, the experiments showed that the proposed method is promising. In the future, we intend to combine RL to advancing the development of evolutionary algorithms.






\section{Appendix}

In this section, we briefly introduce the concepts used in reinforcement learning. Basically, RL aims to maximize the expected cumulative reward, i.e. $\mathbb{E}\left(\sum_{t=1}^T \gamma^t r_t \right)$. First define
\begin{align*}
G_t=r_{t+1}+\gamma r_{t+2}+\cdots+\gamma^{T-t-1}r_T
\end{align*}The expectation of $G_t$ measures the benefit on time $t$. Further, we define the state-value function $v(s)$ and action-value function $Q(s,a)$ as follows:
\begin{align}
v(s)&\triangleq\mathbb {E}[G_t|s_t=s]\\
Q(s,a)&\triangleq\mathbb{E}[G_t|a_t=a,s_t=s]
\end{align}
Without loss of generality, set $\gamma=1$, we have the following Bellman's equality:
\begin{align*}
v(s)&=\mathbb {E}[r_{t+1}|s_t=s]\\&+ \mathbb{E}[\mathbb{E}[r_{t+2}+\cdots+r_T|s_{t+1}=s']|s_t=s]\\
&=\mathbb {E}[r_{t+1}+v(s')|s_t=s]
\end{align*}For action value function, we also have the Bellman's formula:
\begin{align*}
Q(s,a)=\mathbb {E}[r_{t+1}+v(s')|s_t=s,a_t=a]
\end{align*}
For optimal policy $\pi(a|s)$, we have:
\begin{align*}
v(s)=\max_a Q(a,s)
\end{align*}which induces the optimal Bellman equation:
\begin{align}
Q(s,a)=\mathbb {E}[r_{t+1}+\max_{a'} Q(a',s')|s_t=s,a_t=a]
\end{align}This resembles line~\ref{q4} of Alg.~\ref{alg:Q_learning}.

\bibliographystyle{IEEEtran}

\bibliography{ref}

\end{document}